%% file: main.tex
\documentclass[10pt,journal,compsoc]{IEEEtran}

\usepackage[utf8]{inputenc} 
\usepackage{hyperref}       
\usepackage{url}            
\usepackage{booktabs}       
\usepackage{amsfonts}       
\usepackage{nicefrac}       
\usepackage{microtype}      
\usepackage{color}
\usepackage{amsmath}
\usepackage{color}
\usepackage[algoruled,boxed,lined]{algorithm2e} 
\usepackage{algorithmicx}
\usepackage{mathtools}
\usepackage{subcaption}
\usepackage{graphicx}
\graphicspath{ {./images/} }

\newcommand{\etal}[1]{\textit{#1 et al.}}
\newcommand{\best}[1]{\underline{#1}}

\ifCLASSOPTIONcompsoc
  \usepackage[nocompress]{cite}
\else
  \usepackage{cite}
\fi
\ifCLASSINFOpdf
\else
\fi
\begin{document}
%

\title{Sharing Matters for Generalization in \\ Deep Metric Learning}
%
%
%
%

\author{Timo~Milbich*, 
        Karsten~Roth*, 
        Biagio~Brattoli, 
        and~Björn~Ommer 
\IEEEcompsocitemizethanks{
\IEEEcompsocthanksitem The authors are with Heidelberg University, Heidelberg Collaboratory for Image Processing (HCI) and IWR, Mathematikon (INF 205), D-69120, Germany.
E-mail: $\{$firstname.surname$\}$@iwr.uni-heidelberg.de
}
\thanks{* Indicates equal contribution.}
}

\IEEEtitleabstractindextext{%
\begin{abstract}
Learning the similarity between images constitutes the foundation for numerous vision tasks. The common paradigm is discriminative metric learning, which seeks an embedding that separates different training classes. However, the main challenge is to learn a metric that not only generalizes from training to novel, but related, test samples. It should also transfer to different object classes. So what complementary information is missed by the discriminative paradigm? Besides finding characteristics that \emph{separate} between classes, we also need them to likely occur in novel categories, which is indicated if they are \emph{shared} across training classes. This work investigates how to learn such characteristics without the need for extra annotations or training data. By formulating our approach as a novel triplet sampling strategy, it can be easily applied on top of recent ranking loss frameworks. Experiments show that, independent of the underlying network architecture and the specific ranking loss, our approach significantly improves performance in deep metric learning, leading to new the state-of-the-art results on various standard benchmark datasets.
\end{abstract}

\begin{IEEEkeywords}
Deep Metric Learning, Generalization, Shared Features, Image Retrieval, Similarity Learning, Deep Learning.
\end{IEEEkeywords}}

\maketitle

\IEEEdisplaynontitleabstractindextext

%
\IEEEpeerreviewmaketitle

\IEEEraisesectionheading{\section{Introduction}\label{sec:intro}}

%
%
%
%

\begin{figure}[t]
  \centering
    \includegraphics[width=9cm]{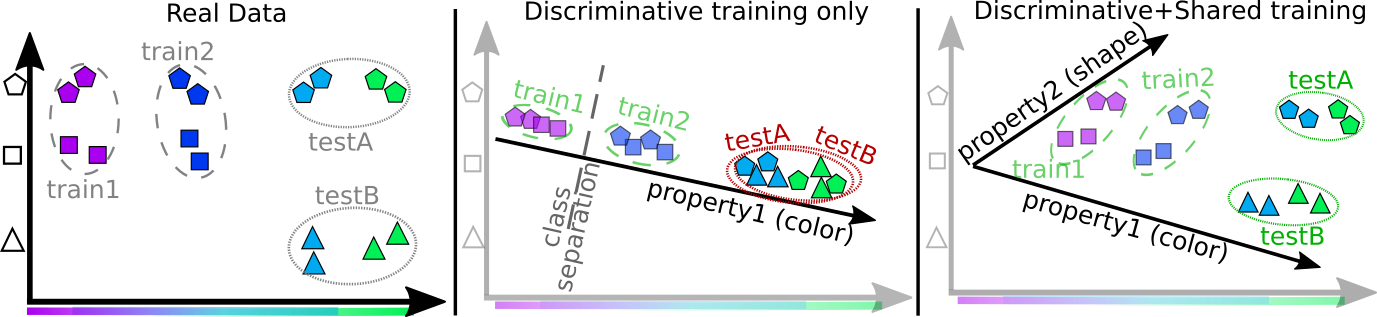} 
    \caption{\textit{Motivation}. (left) Real data is described by several latent characteristics (color and shape). (center) Only the discriminative characteristics (color) which separate classes are learned, while the others are ignored, thus failing to generalize to unseen classes. (right) Including characteristics shared across classes (shape) leads to a better representation of the data. }
    \label{fig:motivation}
\end{figure}

\IEEEPARstart{L}{earning} visual similarities is essential for a wide variety of applications in computer vision, e.g. image retrieval\cite{proxynca,dvml,margin}, zero-shot learning\cite{lifted,facilitylocation}, human pose estimation\cite{cliquecnn,milbich2017unsupervised,brattoli2017lstm,lorenz2019unsupervised} or face verification\cite{semihard,npairs}. Deep metric learning (DML)\cite{semihard,margin,npairs} is currently the main paradigm for learning similarities between images. A deep neural network learns an embedding space which maps related images onto nearby encoding vectors and unrelated ones far apart. The main challenge is then not just maximizing generalization performance from a training to an independent and identically distributed test set. Rather, DML typically aims at transfer learning, i.e., discovering an embedding that is also applicable for differently distributed test data. A typical example is training and test data that exhibits entirely different classes. This degree of generalization is significantly more demanding. It requires to learn visual characteristics that generalize well and likely transfer to unknown object classes. 
\\
Current DML approaches are mostly trained using variants of triplet losses~\cite{semihard,margin,npairs}.
Two samples of a triplet, the anchor and positive, are pulled together in the embedding space, while pushing away a third one, which acts as a negative. 
The task is then typically framed as learning only the characteristics which \emph{separate} the classes while being invariant to all those \textit{shared across} classes. The underlying assumption is that features that discriminate between training classes will also help to separate between arbitrary other test classes. However, as these features accurately circumscribe each training class, it is unlikely that they will generalize to novel classes, cf. Fig. \ref{fig:motivation}. Therefore, to learn a metric that generalizes, we need to find a complementary source of features in our data.

\noindent
Additionally to the class-specific discriminative features, we propose to explicitly learn those so far neglected characteristics that are shared across samples of various training classes. Since such features are of more general nature, they are more likely to generalize to unseen test classes. Therefore we develop a strategy to learn such shared features without the need for extra supervision or additional training data. Our approach is formulated as a novel triplet sampling strategy which explicitly leverages triplets connecting images from mutually different classes. Consequently, it can be easily employed by any ranking loss framework. 
\\
Our main contributions which extend the idea of our earlier work~\cite{mic} are summarized as follows: (i) We introduce the concept of shared characteristics into DML on a more general level and analyze its importance for successful DML generalization. To this end, (ii) we examine how standard discriminative approaches suffer from impaired generalization capabilities and show how shared features help to alleviate these issues while providing a complementary training signal; (iii) we propose a novel and simple method to effectively learn shared characteristics without the need for extra data or annotations and, further, overcome the shortcomings of our previous heuristic-based approach; (iv) we present an effective strategy for incorporating the learning of shared characteristics into classic discriminative ranking loss frameworks, thereby strongly boosting generalization performance.
\\
Experiments using different ranking losses and architectures on standard DML benchmarks show consistent improvements over the state-of-the-art. We further investigate our model and its generalization ability using ablation studies.

\begin{figure*}[t]
  \centering
  \includegraphics[width=0.99\textwidth]{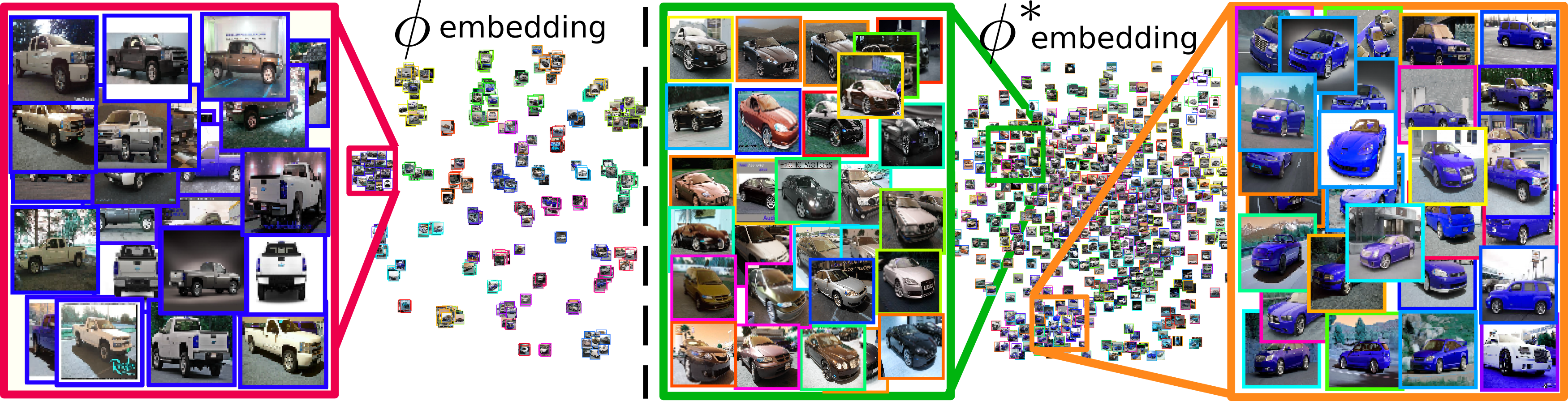}
  \caption{\textit{tSNE\cite{tsne} projections of encoding spaces.} (left) Discriminative training of embedding space on $\mathcal{T}_{\mathcal{X}}$ only, (right) Shared training of embedding space on $\mathcal{T}_{\mathcal{X}}^*$ only. The same random image subset from the CARS196\cite{cars196} training set is visualized. Image contour color indicates ground-truth class. (left) the embedding groups images into compact class-based clusters, (right) the embedding aligns images based on characteristics shared across classes, e.g. view point (green) and color (orange), which are likely to generalize. See supplementary for larger version.}
  \label{fig:tsne}
\end{figure*}

\section{Related Work}\label{sec:related}
\textbf{Ranking losses:} Triplet loss\cite{semihard} has become the standard for deep metric learning in recent years. Several works have proposed methods that go beyond triplets, searching for a more global structure\cite{lifted,npairs,histogram,facilitylocation,angular,rankedlist,proxynca}. For example, \etal{Song}\cite{lifted} proposed a loss to exploit all negatives in the batch. ProxyNCA\cite{proxynca} optimizes with reference to class proxies, thus reducing the complexity problem. Most of the research in the field, however, focuses on optimizing the efficiency of triplets sampling, e.g. by selecting informative negatives\cite{semihard,margin,smartmining,htg} or generating them\cite{daml,dvml,hardness-aware}. Most effectively, \etal{Wu}\cite{margin} propose to sample negatives uniformly from the whole range of possible distances to the anchor. \etal{Zheng}\cite{hardness-aware} uses linear interpolation to produce negatives closer to the anchor. 
\\
\\
\textbf{Ensemble methods:} Another line of research is searching for an effective way to combine multiple encodings in ensembles\cite{abier,hdc,Sanakoyeu_2019_CVPR}. \etal{Opitz}\cite{abier} trains many encoding spaces using the same discriminative task while reducing their mutual information. DREML\cite{dreml} partitions the label space into subsets by means of whole classes and learns independent encoders optimizing the same standard label-based DML task. In contrast, our approach does not represent an ensemble method, as we optimize dedicated encoders for inherently different tasks: standard discriminative DML \textit{and} learning complementary shared characteristics. A novel line of research has been introduced by \etal{Lin}\cite{dvml} which proposes to explicitly learn the intra-class variance by training a generative model in parallel to the embedding space. Differently from \cite{dvml}, we directly learn characteristics shared across classes and not only the distribution over the classes. Further, we do not have to revert to a costly generative model, but use standard triplet formulations which can be naturally integrated into standard ranking loss frameworks.
\\
\\
\textbf{Multi-task learning:}
The general topic of multi-task learning\cite{mtl_bhattarai,mtl_pu} aims at simultaneously learning multiple classifiers on different semantic concepts and/or datasets, thus sharing a similar motivation with our approach. However, compared to Battarai et al.\cite{mtl_bhattarai} we require no additional training data and costly extra annotations as we learn shared features solely from the training data already used for the discriminative task. Pu et al.\cite{mtl_pu} train individual classifiers for groups of whole categories. Our concept of shared features operates on individual samples. Hence, we are able to learn features which are only shared between \textit{some samples} of different classes which is much more flexible.

\section{Approach}\label{sec:method}
In this section, we propose a method to extend the generalization ability of existing metric learning frameworks by incorporating features that are shared across classes. After defining classical discriminative metric learning, we discuss the rationale behind shared features and how to learn them. Finally, we present how to best incorporate discriminative and shared features in one model.

\subsection{Discriminative metric learning} 
\label{sec:disc_dml}
Let $f_i := f(x_i, \theta) \in \mathbb{R}^N$ be a $N$-dimensional feature representation of a  datapoint $x_i \in \mathcal{X}$ parametrized by $\theta$. The objective is to find a consecutive mapping $\phi: \mathbb{R}^N \rightarrow \Phi \subseteq \mathbb{R}^D$ with $\phi_i := \phi(f_i)$, such that similar datapoints $x_i, x_j$ are close in the embedding space $\Phi$ under a predefined distance function $d = d(\phi_i, \phi_j)$ and far from each other if they are dissimilar. Typically $d$ is the euclidean distance and we define $d_{ij} := \left\Vert \phi_i -  \phi_j \right\Vert_2$. Moreover, $f(x_i, \theta)$ is represented as the output of a feature extractor network and $\phi$ is realized as an encoder network (typically a subsequent fully connected layer) with its output normalized to a unit hypersphere $\mathbb{S}^{D}$ (i.e. $\left\Vert \phi_i \right\Vert_2 = 1$) for regularization purposes \cite{semihard}.
\\
The most widely adopted class of training objectives for learning $\phi$ are ranking losses \cite{semihard,margin,rankedlist,icml20}, with the triplet loss\cite{semihard} being its most prominent representative. It is formulated on triplets of datapoints $t = \{x_i, x_j, x_k\}$  as

\begin{equation}
\mathcal{L}(t;\phi) = \mathcal{L}(x_i, x_j, x_k; \phi) = \text{max}(0, d_{ij}^2 - d_{ik}^2 + \alpha)
\label{eq:triplet}
\end{equation}

where $x_i$ is called the anchor, $x_j$ the positive and $x_k$ the negative sample. In supervised metric learning, $x_i$ and $x_j$ are sampled from the same class (i.e. $y_{ij} = 1$) and $x_k$ from another class (i.e. $y_{ik} = 0$). Thus, by optimization on the set of triplets $\mathcal{T}_{\mathcal{X}} := \big\{ \{x_i, x_j, x_k\} \in \mathcal{X}^3 : \; y_{ij} = 1 \wedge y_{ik} = 0 \big\}$, metric learning is framed as a discriminative task. Intuitively $\mathcal{L}(t;\phi)$ imposes a relative ordering on the distances within $t$, pushing $x_i$ closer to $x_j$ than $x_k$ by at least a fixed margin $\alpha$. 
\\
While variants \cite{margin,proxynca,npairs} of the triplet loss have been successfully combined with different triplet sampling strategies \cite{semihard,margin,daml,hardness-aware}, the underlying paradigm of strictly following the user provided class labels imposed on the training data remains unchanged. 


\subsection{Shared characteristics for improved generalization of DML}
\label{sec:disc_lat_prop}
In DML, ranking losses, such as Eq.~\ref{eq:triplet}, enforce mutual similarity of samples from the same training class and dissimilarity to others. Provided a large number of training classes, each class will be accurately separated from all others. This contracts training categories in the embedding space and separates them from one another, as can be seen in Fig.~\ref{fig:motivation}. While such accurate models are beneficial for generalizing from training to i.i.d. test data, transfer to entirely novel classes is significantly more challenging and asks for additional information.
Now, which complementary, so far unused information can we exploit without reverting to additional training data or extra annotations? The class-discriminative task ideally seeks commonalities shared by all samples in a class that separate them from the other classes. A representation that generalizes to different, unknown classes calls for features that are not just discriminative. They should also be shared by different training classes so they are likely to transfer to and reoccur in unknown classes. However, simply merging classes and subsequently learning to separate these super-classes would not be complementary to the existing representation from the class-discriminative task. Moreover, individual classes already have a large intra-class variability. Learning a common representation for even more heterogeneous super-classes would, therefore, be only more difficult and prone to noise. In contrast, we could learn complementary features that are shared across \emph{subsets of different classes} while still separating from other subsets. Such a representation would be \emph{(i)} complementary to the class-discriminative one, since discrimination is between subsets of the original classes and \emph{(ii)} more likely to transfer, since its features are already shared across classes. However, this directly raises the question of finding such subsets without supervision on what is shared between which subsets. We will now first discuss a grouping-based approach for learning shared features between the training samples in $\mathcal{X}$ before presenting a more direct and simple solution.

\begin{figure*}[t]
  \centering
  \includegraphics[width=0.99\textwidth]{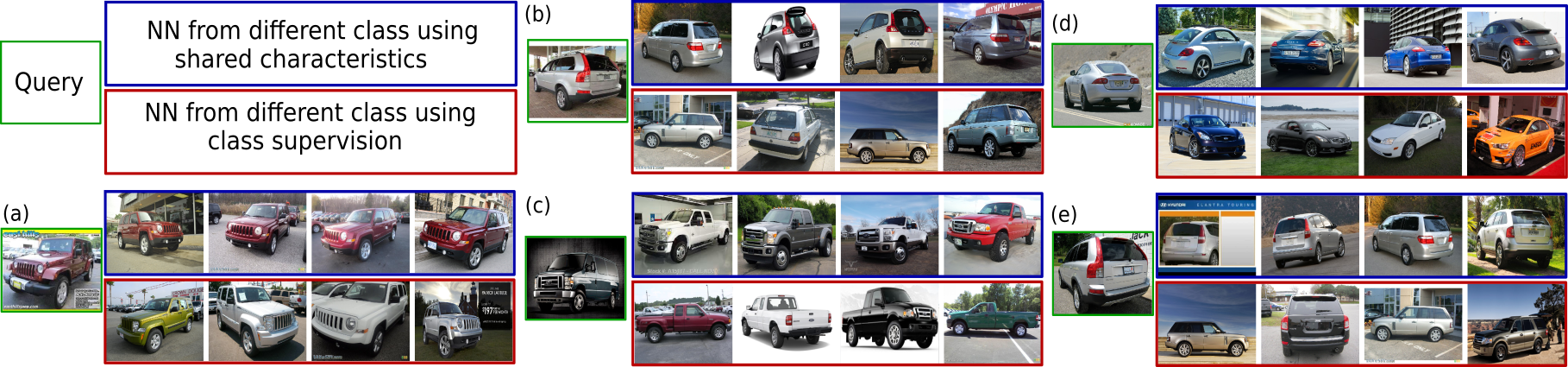}
  \caption{\textit{Nearest neighbour retrieval using $\phi$ and $\phi^*$}. Based on the class- ($\phi$) and shared ($\phi^*$) embedding, we show nearest neighbor retrievals limited to images with different class label than the query image. The neighbors obtained based on the embedding $\phi^*$ trained for shared characteristics exhibit common visual properties. Most prominent: \textit{(a)} red color, \textit{(b)} and \textit{(c)} pose and car type, \textit{(d)} roundish shape, \textit{(e)} back view and color. The embedding $\phi$ trained for class discrimination fails to consistently retrieve meaningful neighbours outside of the query's class.}
  \label{fig:nn}
\end{figure*}

\begin{figure}[b]
  \centering
  \includegraphics[width=9cm]{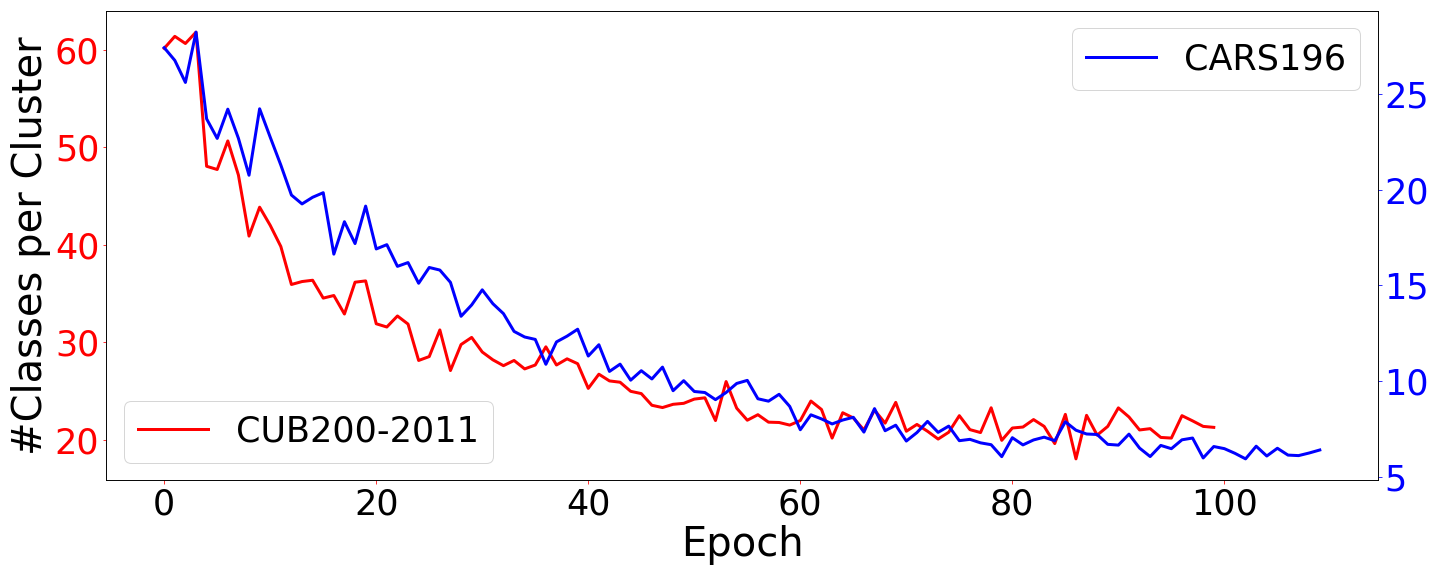}
  \caption{\textit{Unique classes per group.} The average number of unique classes per group decreases during training on CARS196\cite{cars196} and CUB200-2011\cite{cub200-2011} dataset.}
  \label{fig:cluster_progression}
\end{figure}

\vspace{3mm}

\noindent
\textbf{Learning shared features by grouping} 
The prevailing learning strategy in absence of supervision is that of clustering-based methods\cite{deepcluster,cliquecnn,pr20_reliable_relations}. Given a feature representation $f$, the training data $\mathcal{X}$ is partitioned into $L$ groups $\mathcal{G}_l$, $l \in {1,\dots,L}$. Mutual closeness of group members due to similar feature representations indicates that these members share certain characteristics. Consequently, discriminating groups $\mathcal{G}_l$ from another encourages the model to learn about these shared group characteristics and the corresponding features that distinguishes different groups. Thus, we can formulate the learning objective as minimization of a standard discriminative ranking loss, such as Eq. \ref{eq:triplet}, on triplets defined based on the assignment of samples $x_i$ to groups $\mathcal{G}_l$~\cite{mic}, i.e. triplets $t^{\mathcal{G}_l} \in \mathcal{T}_{\mathcal{X}}^{\mathcal{G}_l} := \big\{ \{x_i, x_j, x_k\} \in \mathcal{X}^3 : \; y_{ij}^{\mathcal{G}_l} = 1 \wedge y_{ik}^{\mathcal{G}_l} = 0 \big\}$. Here $y_{ij}^{\mathcal{G}_l} = 1$ denotes samples $x_i,x_j$ belonging to the same group $\mathcal{G}_l$ and $y_{ij}^{\mathcal{G}_l} = 0$ indicates $x_i,x_j$ to come from different groups $\mathcal{G}_l,\mathcal{G}_m, l \neq m$.

\begin{figure*}[t]
  \centering
    \includegraphics[width=0.99\textwidth]{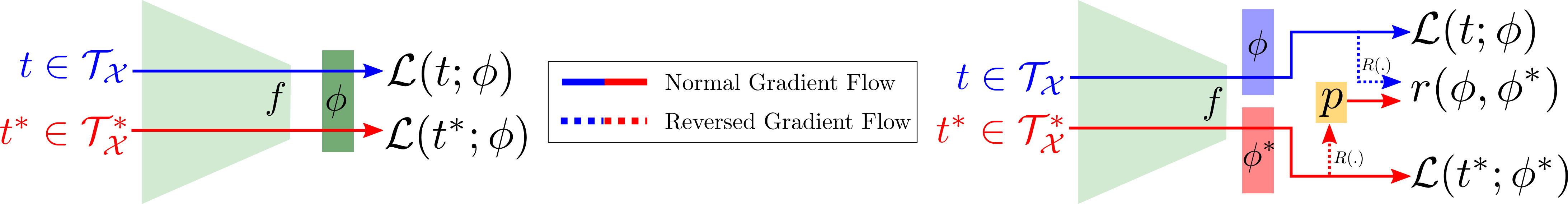}
    \caption{\textit{Architecture and gradient flow.} (left) A single encoder $\phi$ is alternately trained on both tasks. (right) Each task is trained on a dedicated encoder $\phi$ and $\phi^*$ based on a shared feature extractor $f$. Loss is computed per encoder and back-propagated through the shared feature representation $f$. Using a projection network $p$, we map $\phi^*$ to the embedding $\phi$ and compute the decorrelation loss (eq.~\ref{eq:correlation}) with gradient reversal $R(.)$.}
    \label{fig:architecture}
\end{figure*}

\noindent
Unfortunately, clustering-based models are typically strongly biased to learn class-specific structures \cite{deepcluster}, since images from the same class share many common (class-)properties and thus are likely to be assigned to similar groups. Consequently, in order to learn features which are complementary to the class-discriminative task, we first have to reduce the influence of class characteristics. For this purpose, we perform a per-class feature standardization before grouping our data $\mathcal{X}$: For each ground-truth class $c \in \mathcal{C}$ we compute the mean $\mu_c$ and the diagonal of the covariance matrix $\sigma_c$ based on the features $f_i$ of samples $x_i \in \mathcal{X}$ belonging to class $c$. To obtain a grouping $\mathcal{G}_l$, we next apply a clustering algorithm like K-Means~\cite{kmeans} on the standardized features $f_i = \frac{f_i - \mu_{c}}{\sigma_{c}}$, thus reducing the impact of class-specfic information on the feature representations before grouping as presented in our earlier work~\cite{mic}.
\\
\\
\textbf{Explicit inter-class triplet constraints.} 
\label{sec:shared_explicit}
The just described procedure enables learning of characteristics shared within a group $\mathcal{G}_l$. However, even though applying feature standardization the impact of class-specific information on these characteristics is still significant as Fig. \ref{fig:cluster_progression} reveals. During training, each group $G_l$ is gradually dominated by only few classes. Consequently, by sampling anchors $x_i$ and positives $x_j$ of triplets $t^{\mathcal{G}_l}$ from the same group ($y_{ij}^{\mathcal{G}_l} = 1$), $x_i, x_j$ are increasingly likely to also have the same class label ($y_{ij} = 1$). As a result, due to a growing intersection between $\mathcal{T}_{\mathcal{X}}$ and $\mathcal{T}_{\mathcal{X}}^{\mathcal{G}_l}$, lots of class-discriminative, thus redundant features are learned. Concluding, only those triplets $t^{\mathcal{G}_l} \in \mathcal{T}_{\mathcal{X}}^{\mathcal{G}_l}$ will actually provide for new, complementary features, where \textit{each} constituent comes from a different ground-truth class so that $x_i$ and $x_j$ are unlikely to share class-specific properties. Following this intuition, we hypothesize that for almost any arbitrarily formed triplet $t^*$ of $x_i,x_j,x_k$ from mutually different classes, the anchor $x_i$  and positive $x_j$ share \textit{some} common pattern when compared to a third, negative image $x_k$. 
\\
Let $t^*$ be such a triplet of images from the set $\mathcal{T}_{\mathcal{X}}^* := \big\{\{x_i,x_j,x_k\} \in \mathcal{X}^3: y_{ij} = y_{ik} = y_{jk} = 0 \big\}$. For each $t^*$, the commonality between $x_i$ and $x_j$ either represents \emph{(i)} actual shared characteristics across classes which are repetitively supported by many other triplets $t^*$ or \emph{(ii)} some unique or rarely occurring pattern which is typically referred to as noise. Learning such informative characteristics while discarding noisy patterns on $\mathcal{T}_{\mathcal{X}}^*$ constitutes the classical task of DML: Due to the nature of stochastic gradient decent training, deep neural networks only learn by being repetitively exposed to similar training signals. Thus, only the most frequently occurring patterns, i.e. shared characteristics, corroborate their signals and are captured during training, e.g. when learning on imbalanced training classes~\cite{imb1}. Moreover, the learned features are guaranteed to be complementary and also discriminate between different shared characteristics since $x_i, x_j, x_k$ are forced to come from different classes. Fig~\ref{fig:tsne} verifies this by comparing the learned embedding spaces by DML trained on $\mathcal{T}_{\mathcal{X}}$ and $\mathcal{T}_{\mathcal{X}}^*$.
\\
\\
\textbf{Online sampling of inter-class triplets $t^*$.} Shared features can basically be learned between any given training classes from corresponding triplets $t^*$. However, due to the common regularization of $\Phi$ to a unit hypersphere with large dimensionality $D$ (cf. Sec. \ref{sec:disc_dml}), distances in $\Phi$ are strongly biased towards the analytical mean distance~\cite{p_hypersphere}. Thus, to learn shared features between classes from the whole range of distances in $\phi$ and, in turn, increase their diversity, we employ distance-based sampling~\cite{margin}: For each anchor $x_i$ in a mini-batch, we sample the triplet constituents from the remaining mini-batch inversely to the analytical distance distribution $q(d) \propto d^{D-2}\left[ 1 - \frac{1}{4}d^2 \right]^{\frac{D-3}{2}}$ (for large $D \geq 128$\cite{p_hypersphere}) of distances $d$ on $\mathbb{S}^{D}$.

\subsection{Deep metric learning by combining shared and discriminative characteristics}
\label{sec:comb_setup}

The complementary features learned in the previous section represent a complementary source of information to the discriminative features from Sec.~\ref{sec:disc_dml}. Thus, to maximize generalization performance both should be combined. The following discusses strategies for integrating both characteristics, which are compared in the experiments of Sec.~\ref{sec:architecture}. 
\\
As we formulated learning shared features as a novel triplet sampling strategy, it can easily be incorporated into any standard ranking loss framework (in the following we use $\mathcal{T}_{\mathcal{X}}^*$). The most natural way to combine both kinds of features is to alternately optimize $\mathcal{L}$ on $t \in \mathcal{T}_{\mathcal{X}}$ and $t^* \in \mathcal{T}_{\mathcal{X}}^*$ using the same encoder $\phi$. The combined loss is then formulated as $\mathcal{L}^c = \mathcal{L}(t;\phi) + \mathcal{L}(t^*;\phi)$.
\\
However, as similarity learned from $t$ and $t^*$ is based on different semantic concepts (class label vs. inter-class characteristics), simultaneous optimization of a joint encoder $\phi$ may diminish the individual training signals of both tasks. Hence, to fully exploit the overall training signal, we optimize a dedicated $d^*$ dimensional embedding space $\Phi^*$ on $t^*$ to capture the shared characteristics. This requires a second encoder $\phi^*: \mathbb{R}^N \rightarrow \Phi^* \subseteq \mathbb{R}^{D^*}$ with $\phi_i^* := \phi^*(f_i)$. Note that both encoders $\phi(f_i)$ and $\phi^*(f_i)$ act on the same feature representation $f_i$. Thus, the training signals from $\mathcal{L}(t;\phi)$ and $\mathcal{L}(t^*; \phi^*)$ are merged into the same feature extractor network by backpropagation. Consequently, even though $\phi$ and $\phi^*$ optimize different embedding spaces, both benefit from learning to represent shared \textit{and} discriminative characteristics in $f_i$. Fig.~\ref{fig:nn} shows that shared characteristics are prominent in $\phi^*$, while $\phi$ is almost random when searching for nearest neighbors across different classes.
\\
While either learning task contributes complimentary information to the joint feature extractor $f$, there may still be redundant overlap in the their training signals. In order to maximize the diversity and information of the overall training signal, we will subsequently decorrelate the embeddings $\phi$ and $\phi^*$. In a first step, we need to make them comparable by learning a projection $p: \mathbb{R}^{D^*} \rightarrow \mathbb{R}^D $ from $\phi_i^*$ to $\phi_i$. This is a regressor network that is trained by maximizing the correlation $r$ between the projection $p(\phi_i^*)$ and $\phi_i$,

\begin{equation}
r(\phi_i, \phi_i^*) = \frac{1}{D} \sum_{s=1}^{D} (\phi_{i,s} \cdot p(\phi_{i}^*)_s)^2
\label{eq:correlation}
\end{equation}

Here $\phi_{i,s}, p(\phi_{i}^*)_s$ denotes the $s$-th dimension of the respective encoding. While the regressor seeks to maximize the correlation $r$, we invert its gradients (and thus the corresponding training signal) to the embedding representations during backpropagation using a gradient reversal $R(.)$ which flips the gradient sign. As a result, we actually learn to minimize the correlation $r$ and, hence, de-correlate $\phi$ and $\phi^*$. This procedure is inspired by \cite{abier}, which optimizes an ensemble of learners on the same discriminative DML task so each learner is active on different training classes. In contrast, we aim at learning separate embedding spaces using different training tasks to capture shared and discriminative characteristics with minimum overlap. This yields our final training objective

\begin{equation}
\mathcal{L}^{\text{c+decor}} = \mathcal{L}(t;\phi) + \mathcal{L}(t^*;\phi^*) - \gamma \cdot r(R(\phi), R(\phi^*))
\label{eq:triplet_two_corr}
\end{equation}

where $r(R(\phi), R(\phi^*))$ denotes the de-correlation between $\phi, \phi^*$ by applying Eq.~\ref{eq:correlation} on the individual embedded triplet constituents and the subsequent gradient inversion $R(.)$. The parameter $\gamma$ balances the metric learning tasks with the de-correlation. 
Fig. \ref{fig:architecture} visualizes gradient flow and training signal of each embedding $\phi, \phi^*$ and Algorithm 1 summarizes the training procedure.
\\
After training, we can now combine the information captured in both encodings by concatenating $\phi_i$ and $\phi_i^*$ and obtain distances $d((\phi_i,\phi_i^*)^\top, (\phi_j,\phi_j^*)^\top)$. However, the experimental analysis in Sec.~\ref{sec:analysis_and_generalization} shows that both encoders individually already improve over standard DML. By effectively exploiting the shared and discriminative information captured by the feature extractor $f$ and reducing the bias towards the training classes, they exhibit increased generalization onto the test distribution.



\input{algorithm.tex}

\section{Experiments and analysis}
This section presents technical details of our implementation followed by an evaluation of our model on standard metric learning benchmarks. Furthermore, we analyze generalization in standard deep metric learning compared to our approach and additionally present ablation studies of our proposed model. 
\\
\\
\noindent
\textbf{Implementation details.} We follow the training protocol of \cite{margin} for ResNet50 and \cite{npairs} for GoogLeNet utilizing the original images without object crops. During training, each image is resized to $256\times 256$, followed by a random crop to $224\times 224$ for ResNet50 and $227\times 227$ for GoogLeNet, as well as random horizontal flipping. For all experiments, learning rates are set to $10^{-5}$ for ResNet50 and $10^{-4}$ for GoogLeNet. We choose the triplet parameters according to \cite{margin}, with $\alpha = 0.2$. For margin and triplet loss with semihard sampling\cite{semihard}, $m = 4$ images are sampled per class until the batch size is reached\cite{margin}. For ProxyNCA\cite{proxynca} $m=1$, and for n-pair loss\cite{npairs} $m=2$. 
The regressor network $p$ (Sec. \ref{sec:comb_setup}) is implemented as a two-layer fully-connected network with ReLU-nonlinearity inbetween. We obtain the weighting parameter $\gamma$ for decorrelation by cross-validation within a range of $[500,2000]$, depending on the dataset. For implementation we use the PyTorch framework\cite{pytorch}. All experiments are performed on a single NVIDIA Titan X. While training, we use the same ranking loss for optimizing $\phi^*$ and $\phi$, except for ProxyNCA\cite{proxynca}. In this case we train $\phi^*$ using triplet loss with semi-hard negative mining\cite{semihard}, as $\mathcal{T}_{\mathcal{X}}^*$ operates on individual triplets and thus the concept of proxies is not applicable. For n-pair loss\cite{npairs} we utilize our described method extended to multiple random negatives.
\\
\\
\noindent
\textbf{Benchmark datasets.}
We evaluated on three standard benchmarks for DML reporting image retrieval performances using Recall@k\cite{recall} and the normalized mutual information score (NMI)\cite{nmi}. Our evaluation protocol follows \cite{margin}. For each dataset, we use the first half of the classes for training and the second half for testing. \textit{CARS196}\cite{cars196}: 16,185 car images divided in 196 classes. \textit{Stanford Online Products (SOP)}\cite{lifted}: 120,053 images in 22,634 classes from 12 product categories. \textit{CUB200-2011}\cite{cub200-2011}: 200 classes of bird species for a total of 11,788 images. 

\subsection{Results and comparison with previous works}

In Tab. \ref{tab:cars}, \ref{tab:cub} and \ref{tab:sop} we compare our approach to state-of-the-art DML methods based on the standard image retrieval task (Recall@K). For our methods, we report the results of the concatenated embedding with a dimensionality of 256. If not stated otherwise, we optimize both the discriminative and shared DML tasks using Margin loss\cite{margin}. Thus we also provide its baseline results using 256 dimensions based on our re-implementation for fair comparison. Further, we evaluate our approach for both options to learn shared features, i.e. leveraging $\mathcal{T}_{\mathcal{X}}^{\mathcal{G}_l}$ or $\mathcal{T}_{\mathcal{X}}^*$, respectively. While both options clearly improve over purely discriminative DML methods, using $\mathcal{T}_{\mathcal{X}}^*$ leads to stronger results in general. We conclude that by using triplet constraints which explicitly link different classes, we learn shared features more effectively due to less class-specific information affecting the optimization of the complementary shared feature task. Consequently, in the remainder of the experimental section, we focus on $\mathcal{T}_{\mathcal{X}}^*$. 

\input{tables_new/eval_cars196.tex}
\input{tables_new/eval_cub200-2011.tex}
\input{tables_new/eval_sop.tex}

\input{tables_new/cub_cars_sop_grid.tex}

We outperform other approaches with comparable embedding space capacities by at least $2.4\%$ on CARS196\cite{cars196}, $2.5\%$ on CUB200\cite{cub200-2011} and $1.0\%$ on SOP\cite{lifted}. Moreover, our model outperforms DREML\cite{dreml}, a large ensemble method, by $\sim5\%$ on CUB200-2011 (Tab.~\ref{tab:cub}) and $\sim1\%$ on CARS196 (Tab.~\ref{tab:cars}). Similar behavior is observed for the clustering task (NMI). The results reported by Ranked list\cite{rankedlist} are computed using the InceptionV2\cite{googlenetv2} architecture and a concatenation of three features layers, totaling 1536 dimensions. Similarly to DREML\cite{dreml}, this significantly increases the capacity of the underlying model, making it not directly comparable. 
\\
In Tab. \ref{tab:grid_resnet} we evaluate our approach based on different ranking losses for optimization: triplet loss with semihard negative sampling\cite{semihard}, n-pair loss\cite{npairs}, ProxyNCA\cite{proxynca} and margin loss with distance sampling\cite{margin}. We compare the re-implemented baselines with and without learning complementary shared features. For completeness we present for our approach results based on the individual embeddings $\phi, \phi^*$ and their concatenation after training our model as described in Sec. \ref{sec:comb_setup}. We show the results using ResNet50\cite{resnet} and present additional results using GoogLeNet\cite{googlenet} in our ablation studies. Our method consistently improves upon the state-of-the-art across all datasets, irrespective of architecture and ranking loss. This clearly indicates the universal benefit of shared feature learning.

\subsection{Generalization and Analysis of the Embeddings}
\label{sec:analysis_and_generalization}


This section analyzes the performance of our embeddings $\phi$ and $\phi^*$. It further compares the generalization capabilities of \emph{(i)} classic deep metric learning that is trained only discriminatively (only a single embedding $\phi$ trained on triplets $\mathcal{T}_{\mathcal{X}}$) using margin loss \cite{margin} with that of \emph{(ii)} also exploiting shared characteristics (by leveraging $\mathcal{T}_\mathcal{X}^*$) as suggested in Sec.~\ref{sec:disc_lat_prop} and \ref{sec:comb_setup}. For the analysis we evaluate different encodings on both train- and testset in Tab. \ref{tab:analysis_generalization}. 
\\
\\
\noindent
\textbf{Performance of embedding spaces.}
Tab. \ref{tab:analysis_generalization} summarizes the performance of the discriminatively trained embedding $\phi$, the embedding $\phi^*$ trained for shared characteristics (only for our approach), $\phi$ with random weight re-initialization after training ($\phi(\mathcal{N})$)\footnote{Note that the weights of $f$ remain trained and are not re-initialized.}, and the feature extractor $f$ on the train and test set of CARS196\cite{cars196} dataset. Comparing the test set results of $\phi$ and $f$ (for both \emph{(i)} and \emph{(ii)}) shows that the embedding $\phi$ performs worse than the feature encoding $f$. Consequently, $\phi$ is not able to effectively use the information captured in the features $f$. Comparing $\phi$ to $\phi(\mathcal{N})$ confirms this, since the randomly re-initialized embedding actually performs better in testing. Moreover, learning shared characteristics improves the features $f$, which clearly demonstrates them being complementary to the standard discriminative training signal. Note that in our approach, \textit{both} embeddings ($\phi$ and $\phi^*$) have equal access to the discriminative \textit{and} shared features captured in $f$. However, we observe that $\phi^*$ performs $4.0\%$ better than $\phi$ on the test set. This indicates that $\phi$ is overfitting to the train classes while $\phi^*$ is able to successfully use both, the strong discriminative features and the more general shared features. Thus, $\phi^*$ generalizes more effectively to \textit{unseen} test classes. The next paragraph now further analyzes this observation.
\\
\\
\noindent
\textbf{Generalization analysis.}
Additionally to the performance of the individual embeddings on train and test set, Tab. \ref{tab:analysis_generalization} further shows their difference, the generalization gap. For \emph{(i)} (Tab. \ref{tab:analysis_generalization} top) we observe a large gap of $-10.8\%$ compared to $-5.8\%$ of $\phi$ in our approach, thus indicating strong overfitting. Indeed, simply randomly re-initializing the weights of the encoder $\phi$ before testing already improves the gap to $-6.1\%$ and increases test performance by $1.5\%$. Computing distances based on $f$ further reduces the generalization gap. For \emph{(ii)}, Tab. \ref{tab:analysis_generalization} bottom not only shows a significant increase in test performance as discussed above but also an improvement in generalization compared to \emph{(i)} due to the additional shared characteristics. The generalization gap is significantly smaller for each encoding. Thus, the benefit of learning shared characteristics for improved generalization in the transfer learning problem addressed by DML seems to be twofold: it not only adds complementary information to the features $f$ but also regularizes training to reduce overfitting to the training classes. Note, that overfitting in transfer learning is of different nature and significantly more challenging to counteract than in standard learning settings with i.i.d. training and testing data. In the latter case this issue can be addressed by training on more data from the underlying training distribution or regularizing the adaptation of a model to the available training samples. However, such techniques can only have small impact in the presence of train-test distribution shifts, as even an ideal representation of the training data may not transfer equally well to a unknown testing distribution. Further, as we are measuring the test performance on \textit{unknown test classes}, we do not know if the overfitting effect would also be as severe when evaluating on an i.i.d. test set (as class-discriminative training is actually the standard way to learn on such data). Therefore, to support our hypothesis, we apply standard regularization techniques to a discriminative DML baseline model.

\input{tables_new/analysis_gen_gap.tex}
\input{tables_new/ablation_generalization_techniques.tex}

\vspace{3mm}

\begin{figure*}
  \centering
  \includegraphics[width=0.9\textwidth]{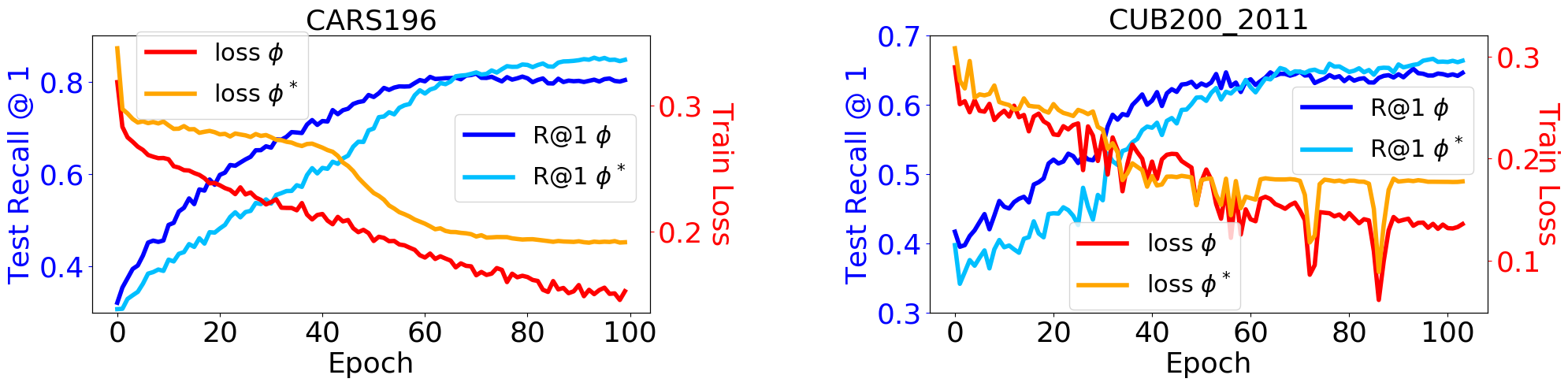}
  \caption{\textit{Training curves.} Train loss and test recall@1 for the class $\phi$ and shared $\phi^*$ encoders. The model is trained using margin loss\cite{margin} with ResNet50\cite{resnet} on CARS\cite{cars196} and CUB\cite{cub200-2011}.}
  \label{fig:performance}
\end{figure*}

\noindent
\textbf{Comparison with standard generalization techniques.}
In Tab.\ref{tab:generalization_noise} we compare our approach with techniques typically used to improve generalization in deep neural networks such as dropout~\cite{dropout} and noise injection~\cite{reg_by_noise}, which proved to be effective for classification and representation learning. As baseline we use margin loss\cite{margin} on ResNet50\cite{resnet}. For fair comparison, we use for our approach the same encoder for both the discriminative and shared task (as also discussed in Tab.~\ref{tab:ablation_architecture} (a) denoted as 'both') and thus the same architecture as for the analyzed techniques. The evaluation is performed on CARS196\cite{cars196} reporting Recall@1 on the test set. In particular, we apply dropout between the features $f$ and the embedding $\phi$ which gives a little boost of $0.6\%$ over the baseline, which is, however, minor respect to the $3.3\%$ gain obtained by our approach. Applying Gaussian noise to the input provides little improvement, while Gaussian noise on the network output reduces performance. Concluding, our approach adds actual complementary information to the discriminative training signal in form of complementary features, and does not only act as a regularization by inducing noise.
\\
\\
\noindent
\textbf{Analysis of training progress}
Fig.~\ref{fig:performance} analyzes the learning behavior of the encoder $\phi$ and $\phi^*$ in our model based on training loss and Recall@1 on the test set during training. Evidently, learning shared characteristics across classes that still separate from others is initially a harder task than only discriminating between classes. Thus, we would expect a weaker performance in earlier training epochs. Further, since shared characteristics are less specialized to the training classes, they yield higher overall performance (cf. Tab.~\ref{tab:analysis_generalization}). And indeed, while $\phi^*$ is initially weaker than $\phi$, it continues to increase when $\phi$ is already saturated.

\vspace{3mm}


\subsection{Ablation studies}
Subsequently, we conduct ablation studies for different aspects of our approach. We first analyze our proposed architecture and then examine different triplet assembling and sampling strategies for learning shared characteristics. Further ablations are shown in the supplementary. \\
\\
\noindent
\textbf{Architecture.}\label{sec:architecture} 
Sec.~\ref{sec:comb_setup} proposes two options for jointly learning discriminative and shared characteristics during training: Alternately optimizing the same single encoding space and learning dedicated embedding spaces for $\phi$ and $\phi^*$ with and without decorrelation. Tab.~\ref{tab:ablation_architecture} (a) compares these options against baselines trained only on either $\mathcal{T}_{\mathcal{X}}$ or $\mathcal{T}_{\mathcal{X}}^*$. Firstly, we observe that only learning the shared task (\textit{shared only}) results in a huge drop in performance to $36.1\%$ compared to $81.5\%$ of the discriminative baseline (\textit{discr. only}). This result is to be expected and easily explained: By neglecting the strong discriminative learning signal obtained from optimizing on $\mathcal{T}_\mathcal{X}$, no class concept is learned, which stands in contrast to the \textit{class-based} evaluation protocol of nearest neighbor retrieval accuracy. This highlights the importance of of the discriminative task for learning a reasonable distance metric. However, \textit{adding} the shared task to the discriminative baseline consistently improves performance independent of our proposed options for joint optimization. Even the simplest option, i.e. using a single encoder for learning from both $\mathcal{T}_{\mathcal{X}}$ and $\mathcal{T}_{\mathcal{X}}^*$ (\textit{both}) improves over the baseline by $1.7\%$. Further, we see an additional gain of $0.9\%$ when optimizing separate encoders, i.e. $\phi$ for $\mathcal{T}_{\mathcal{X}}$ and $\phi^*$ for $\mathcal{T}_{\mathcal{X}}^*$ (\textit{both+sep}) which is explained by the reduced interference between both tasks during training. Finally, using separate encoders allows for explicit de-correlation \textit{(both+sep+decor)} to minimize the overlap in the captured characteristics between $\phi$ and $\phi^*$, yielding another significant gain of $3\%$.
\input{tables_new/ablation_architecture_v2.tex}
\input{tables_new/ablation_sampling_shared_v2.tex}

\input{tables_new/eval_dml_base_inceptionBN.tex}

\vspace{3mm}

\begin{figure*}[h!]
  \centering
  \includegraphics[width=0.93\textwidth]{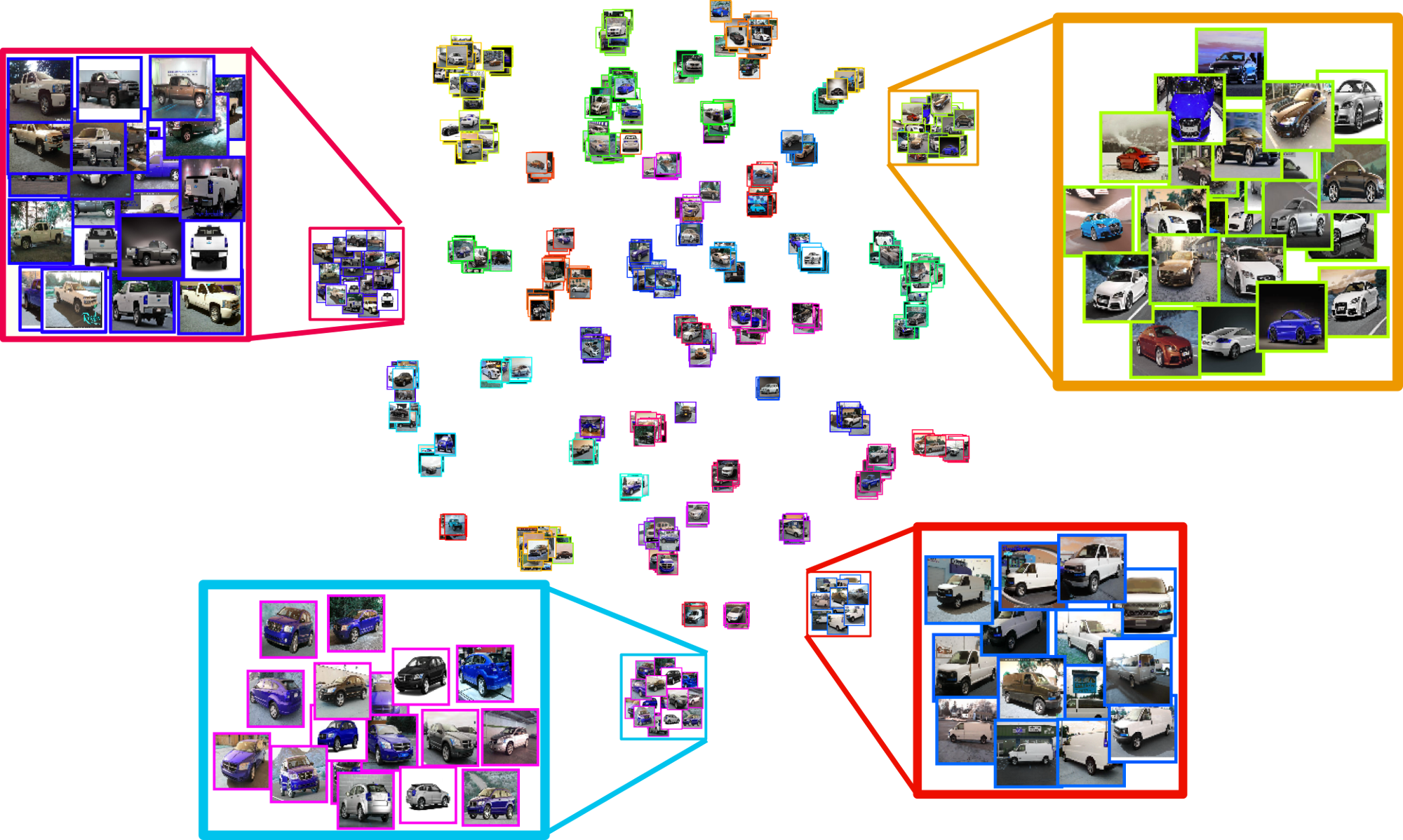}
  \caption{\textit{Embedding space resulting from model trained on discriminative task.} The contour color of the individual images indicate the ground-truth class.}
  \label{fig:tsne_class}
\end{figure*}

\begin{figure*}[h!]
  \centering
  \includegraphics[width=0.93\textwidth]{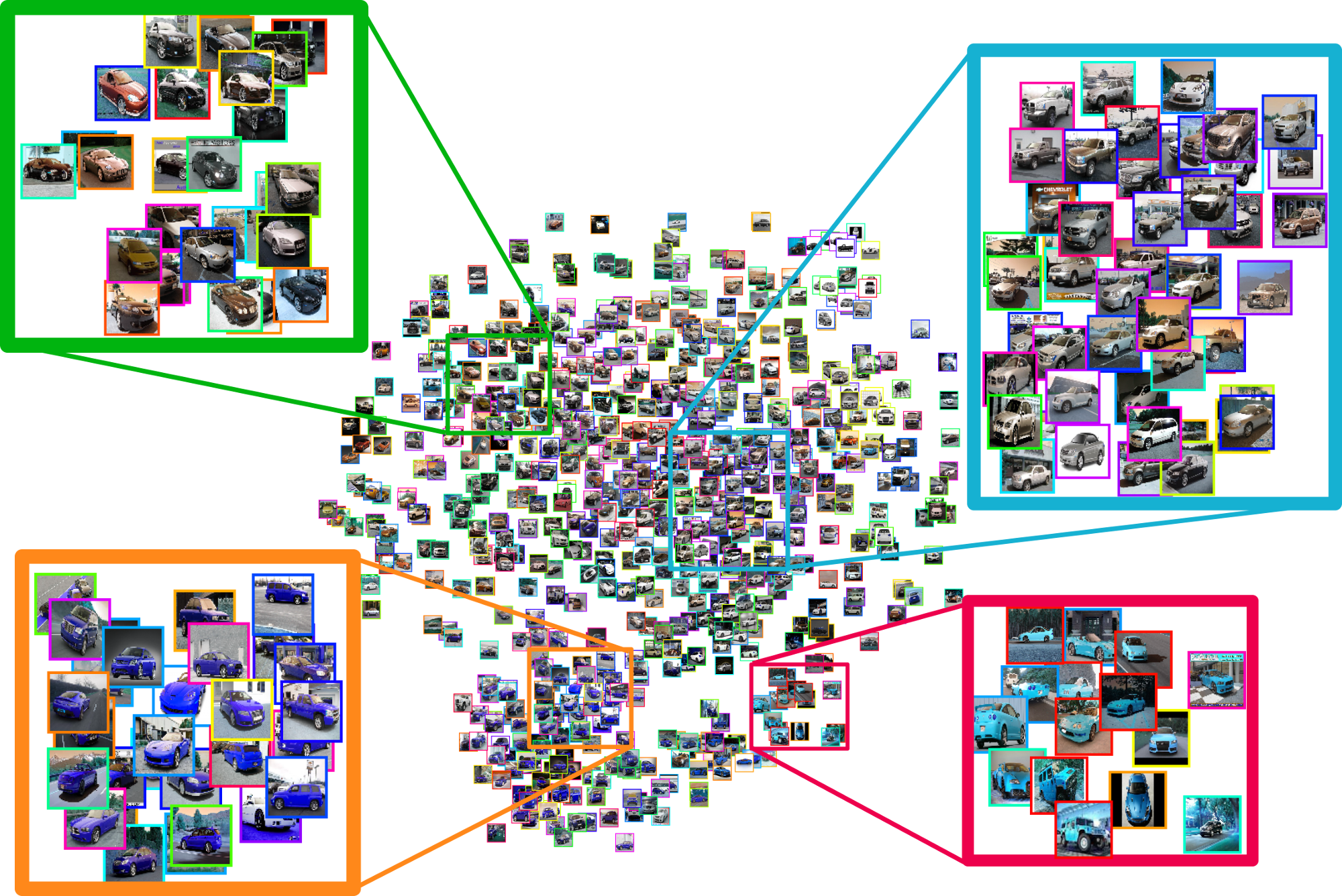}
  \caption{\textit{Embedding space resulting from model trained on shared task.} The contour color of the individual images indicate the ground-truth class.}
  \label{fig:tsne_shared}
\end{figure*}

\noindent
\textbf{Strategies for assembling shared triplets.} Tab.~\ref{tab:ablation_sampling} (a) evaluates different strategies to assemble triplets for learning shared characteristics: \textit{$\mathcal{T}_{\mathcal{X}}^{\mathcal{G}_l}$-std.}: Sampling from $\mathcal{T}_{\mathcal{X}}^{\mathcal{G}_l}$ without using feature standardization before grouping. This strategy adds $0.8$\% to the purely discriminative baseline. However, since these surrogate classes tend to resemble the ground-truth classes in the train set, the performance is significantly worse than our best strategy. Thus, redundant and mostly discriminative signals are added. \textit{$\mathcal{T}_{\mathcal{X}}^{\mathcal{G}_l}$+std.}: Sampling from $\mathcal{T}_{\mathcal{X}}^{\mathcal{G}_l}$ with feature standardization as proposed in \cite{mic}. Even though the effect of class-specific information is strongly reduced and thus performance is boosted by $3.5\%$, this strategy is still inferior to sampling triplets from $\mathcal{T}_{\mathcal{X}}^*$. \textit{min d(a,p)}: Sampling from $\mathcal{T}_{\mathcal{X}}^*$, but for a given anchor, we constrain sampling the positive by always choosing the closest sample in a batch based on distances defined by $\phi^*$. This follows the intuition that mutually close samples are more likely to share some characteristic. Applying this strategy loses $5$\% compared to our best result. We conclude that shared characteristics can be learned from almost all image pairs. Thus, restricting the sample range, neglects very important information. \textit{No constraint}: we report numbers for unconstrained assembling of shared triplets as a proxy for $\mathcal{T}_{\mathcal{X}}^*$, i.e. anchors, positives and negatives are randomly sampled. With $86.0\%$ this simplest strategy works very well. We reason that the performance drop of $1.0\%$ is explained by direct disagreement between some of the triplets sampled from $\mathcal{T}_{\mathcal{X}}^*$ and $\mathcal{T}_{\mathcal{X}}$, distorting the feature extractor $f$. \textit{$\mathcal{T}_{\mathcal{X}}^*$}: Our proposed strategy of sampling each constituent of a triplet from a different class, i.e. sampling triplets from $\mathcal{T}_{\mathcal{X}}^*$.\\
Note that all proposed strategies show improvement over the baseline (margin loss \cite{margin}), clearly proving that learning complementary features is crucial for improved generalization of deep metric learning.
\\
\\
\noindent
\textbf{Influence of de-correlation on generalization.}
To evaluate the relevance of de-correlation between our embeddings $\phi$ and $\phi^*$, Tab. \ref{tab:ablation_architecture}(b) examines generalization performance for different values of $\gamma$. As we see, increasing the de-correlation boosts performance over the non-decorrelated baseline (with $\gamma=0$) for a robust interval. However, if $\gamma$ becomes to large, performance drops below the baseline as enforcing de-correlation strongly dominates the actual DML training signal.
\\
\\
\noindent
\textbf{Strategies for sampling shared triplets.}
In Tab. \ref{tab:ablation_sampling}(b) we investigate different procedures for online sampling of shared triplets $\mathcal{T}_{\mathcal{X}}^*$. We fix the training procedure for $\phi$ (Margin \cite{margin}) and vary the shared triplet sampling procedure. We see that sampling triplets $t^*$ from a broad range of distances and thus diverse classes (cf. Sec.~\ref{sec:disc_lat_prop}) is essential for effectively learning inter-class features. This holds especially for distance-based sampling \cite{margin} which encourages triplets with anchor-negative pairs sampled uniformly over distances.
\\
\\
\noindent
\textbf{Evaluations using GoogLeNet architecture.}
Similar to Tab.~\ref{tab:grid_resnet}, we now present evaluations using the GoogLeNet architecture in Tab.~\ref{tab:grid_ibn}. We provide the results of our re-implementations of the baseline models\cite{semihard,npairs,proxynca,margin} without and in combination with our proposed approach. The experiment shows that, similar to our ResNet50\cite{resnet} results, our approach consistently boosts the baseline models in both image retrieval (Recall@1) and clustering (NMI). In particular, we outperform the best baseline recall by $7\%$ on CARS196, $3\%$ on CUB200-2011 and $3.5\%$ on SOP.
\\
\\
\noindent
\textbf{tSNE projection of embedding spaces}
Fig.~\ref{fig:tsne_class} and \ref{fig:tsne_shared} show $1000$ images sampled randomly from the CARS196\cite{cars196} training set and projected in 2D by applying t-SNE\cite{tsne} on the image encodings. For Fig. \ref{fig:tsne_class} the underlying model is trained solely on the discriminative task, while in Fig \ref{fig:tsne_shared} the model is trained solely on the shared feature tasks. In Fig. \ref{fig:tsne_class} the model learns to group classes very compactly and far from each other, which indicates strong adaptation to the training classes. In Fig. \ref{fig:tsne_shared} images which share visual properties are grouped closer together, independently from their ground-truth labels. This results in complementary features which generalize better to new data. 



\section{Conclusion}\label{sec:conc}
This work has addressed and analyzed the generalization issues of standard deep metric learning approaches arising from their purely discriminative nature. As a remedy, we propose to additionally learn characteristics shared across different classes, which are more likely to transfer to unseen test data. To this end, we additionally train a dedicated encoder on a novel ranking task, explicitly linking samples across classes. Moreover, we show how to combine both discriminative and shared feature learning during training. Evaluations on standard metric learning datasets show that our simple method provides a strong, loss- and architecture independent boost, achieving new state-of-the-art results.

%

\ifCLASSOPTIONcompsoc
  \section*{Acknowledgments}
\else
  \section*{Acknowledgment}
\fi

This work has been supported in part by Bayer AG, the German federal ministry BMWi within the project “KI Absicherung”, and a hardware donation from NVIDIA corporation.

\ifCLASSOPTIONcaptionsoff
  \newpage
\fi




%

%

{\small
\bibliography{bibliography}
}

%

\begin{IEEEbiography}[{\includegraphics[width=1in,height=1.25in,clip,keepaspectratio]{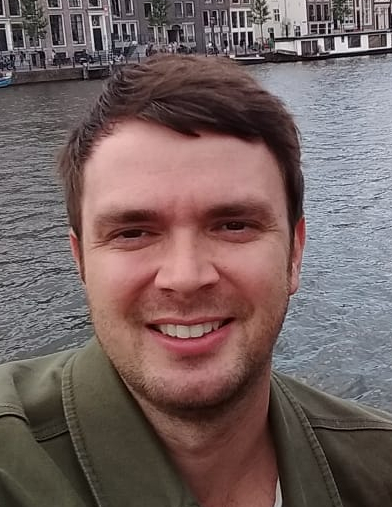}}]{Timo Milbich}
received his masters degree in Scientific Computing
from Ruprecht-Karls-University, Heidelberg, in 2014. He
is currently a Ph.D. Candidate in Heidelberg Collaboratory for
Image Processing at Heidelberg University. His current research
interests include computer vision and machine learning with focus on representation and deep metric learning, as well as human pose understanding. 
\end{IEEEbiography}
\begin{IEEEbiography}[{\includegraphics[width=1in,height=1.25in,clip,keepaspectratio]{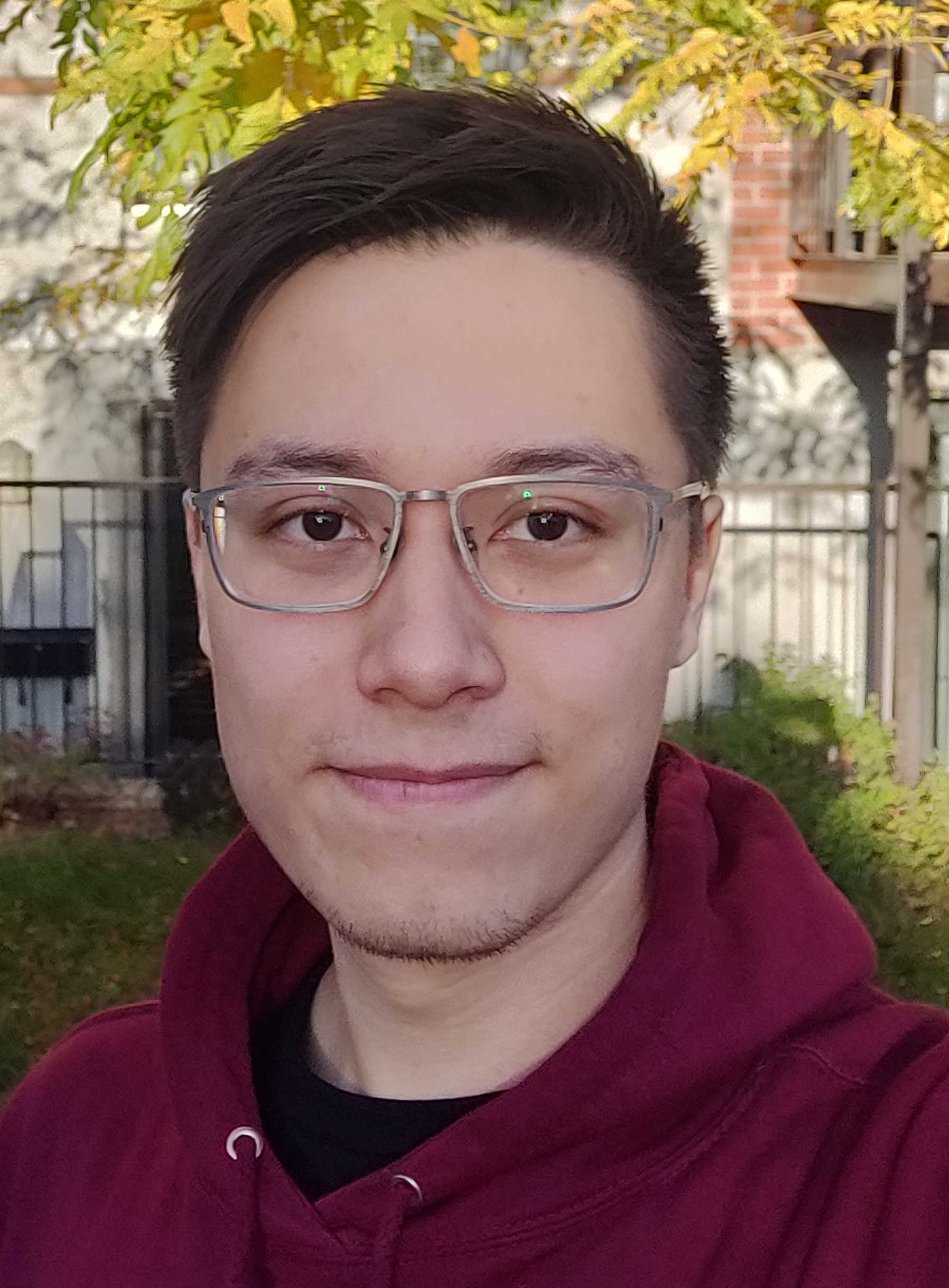}}]{Karsten Roth}
received his bachelors degree in Physics
from Ruprecht-Karls-University, Heidelberg, in 2017. He
is currently a master student in Heidelberg Collaboratory for
Image Processing at Heidelberg University. His current research
interests include computer vision focusing on deep representation and metric learning.
\end{IEEEbiography}
\begin{IEEEbiography}[{\includegraphics[width=1in,height=1.25in,clip,keepaspectratio]{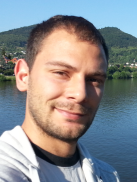}}]{Biagio Brattoli}
received a masted degree in Computer Engineering from Politecnico di Torino, Italy, and in Computer Science from Royal Institute of Technology, Sweden. He is currently a Ph.D. Candidate in Heidelberg Collaboratory for Image Processing at Heidelberg University. His research interests include computer vision and deep learning, focusing on representation learning for images and videos in condition of limited supervision.
\end{IEEEbiography}
\begin{IEEEbiography}[{\includegraphics[width=1in,height=1.25in,clip,keepaspectratio]{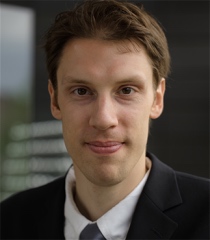}}]{Björn Ommer}
received a diploma in computer science from the University of Bonn, Germany and Ph.D. in computer science from ETH Zurich. Thereafter, he held a postdoctoral position at the University of California at Berkeley. Since 2009 he is heading the computer vision group at Heidelberg University and is a professor with the Department of Mathematics and Computer Science. His research interests include computer vision, machine learning, and cognitive science. He is an associate editor of the IEEE Transactions on Pattern Analysis and Machine Intelligence.
\end{IEEEbiography}
\vfill




\end{document}

%% file: algorithm.tex
\begin{algorithm}[h!]
\label{alg:algorithm}
\caption{Joint training of $\phi$ and $\phi^*$}

\SetKwFunction{Backward}{Backward}
\SetKwFunction{GetBatch}{GetBatch}
\SetKwInOut{Input}{input}
\SetKwInOut{Init}{initialization}
\SetKwInOut{Constants}{parameters}
\SetKwInOut{Output}{output}
\SetKwRepeat{Repeat}{repeat}{until}

\SetAlgoLined

\textbf{Input:}\newline 
Images $\mathcal{X}$, Feature extractor network $f$, Embedding network $\phi$, Embedding network $\phi^*$, Ranking loss $\mathcal{L}$, Batchsize $b$, Decorrelation weight $\gamma$
\newline
    


$epoch$ $\leftarrow$ 0

\While{Not Converged}{
    \Repeat{end of epoch}{
        \emph{Get batches of batchsize b} \\
        $\mathcal{B}$ $\leftarrow$ \GetBatch{$\mathcal{X}$, $b$}\\
        $\mathcal{B}^*$ $\leftarrow$ \GetBatch{$\mathcal{X}$, $b$}
        \newline

        \emph{Sample discriminative triplet for each anchor $x_i$ in $\mathcal{B}$ (cf. Sec. 3.1) based on $q^{-1}(d)$} \\
        $\mathcal{T}_\mathcal{X} \leftarrow \{x_i, x_j, x_k  \stackrel{q^{-1}}{\sim} \mathcal{B} \; | \;  y_{ij}=1 \wedge y_{ik}=0$ \}
        \newline
        
        \emph{Sample explicit inter-class triplet for each anchor $x_i$ in $\mathcal{B}^*$ (cf. Sec. 3.2) based on $q^{-1}(d)$} \\
        $\mathcal{T}_\mathcal{X}^* \leftarrow \{x_i, x_j, x_k  \stackrel{q^{-1}}{\sim} \mathcal{B}^* \; | \;  y_{ij} = y_{ik} = y_{jk}=0$ \} \newline
        
        \emph{Alternate optimization (cf. Sec. 3.3)} \\
        $\ell_\text{discr} \leftarrow$ $\mathcal{L}(\mathcal{T}_\mathcal{X}, \phi) - \gamma\cdot r(R(\phi), R(\phi^*))$ \\
        $\phi, \phi^*, f \leftarrow$ \Backward{$\ell_\text{discr}$}\newline
        
        $\ell_\text{inter} \leftarrow$ $\mathcal{L}(\mathcal{T}^*_{\mathcal{X}}, \phi^*) - \gamma\cdot r(R(\phi), R(\phi^*))$ \\
        $\phi, \phi^*, f \leftarrow$ \Backward{$\ell_\text{inter}$}\newline        
    }

    $epoch \leftarrow epoch+1$ 
}

\end{algorithm}




%% file: tables_new/eval_cars196.tex
\begin{table}[h!]
    \caption{Evaluation on CARS196\cite{cars196}.}
    \label{tab:cars}
    \setlength\tabcolsep{1.5pt}
    \centering
     \begin{tabular}{l|c|ccc|c}
        \toprule
        Approach & Dim & R@1 & R@2 & R@4  & NMI \\
        \midrule
        Rank\cite{rankedlist}[Iv2] & 512 & 74.0 & 83.6 & 90.1 & 65.4 \\
        HTG\cite{htg} & - & 76.5 & 84.7 & 90.4 & - \\
        HDML\cite{hardness-aware} & 512 & 79.1 & 87.1 & 92.1 & 69.7 \\
        Margin\cite{margin} & 128 & 79.6 & 86.5 & 90.1 & 69.1 \\
        HTL\cite{htl} & 512 & 81.4 & 88.0 & 92.7  & - \\
        DVML\cite{dvml} & 512 & 82.0 & 88.4 & 93.3 & 67.6 \\
        MIC\cite{mic} & 128 & 82.6 & 89.1 & 93.2 & 68.4\\
        D\&C\cite{Sanakoyeu_2019_CVPR} & 128 & 84.6 & 90.7 & 94.1 & 70.3\\
        \hline
        A-BIER\cite{abier} & 512 & 82.0 & 89.0 & 93.2  & - \\
        Rank\cite{rankedlist}[Iv2] & 1536 & 82.1 & 89.3 & 93.7& 71.8 \\
        DREML\cite{dreml} & 9216 & 86.0 & 91.7 & 95.0 & 76.4 \\
        \hline
        Margin(ReImp) & 256 & 81.5 & 88.1 & 92.8 & 67.4 \\
        Ours ($\mathcal{T}_{\mathcal{X}}^{\mathcal{G}_l}$) & 256 & 85.8 & 91.3 & 95.1 & \textbf{70.3} \\
        \textbf{Ours ($\mathcal{T}_{\mathcal{X}}^*$)}\cite{mic} & 256 & \textbf{87.0} & \textbf{92.1} & \textbf{95.4} & 69.8 \\
        \bottomrule
    \end{tabular}
\end{table}

%% file: tables_new/eval_cub200-2011.tex
\begin{table}[h!]
    \caption{Evaluation on CUB200-2011\cite{cub200-2011}.}
      \label{tab:cub}
      \setlength\tabcolsep{1.5pt}
      \centering
      \begin{tabular}{l|c|ccc|c}
        \toprule
        Approach & Dim & R@1 & R@2 & R@4 &  NMI \\
        \midrule
        DVML\cite{dvml} & 512 & 52.7 & 65.1 & 75.5 &61.4 \\
        HDML\cite{hardness-aware} & 512 & 53.7 & 65.7 & 76.7 & 62.6 \\
        HTL\cite{htl} & 512 & 57.1 & 68.8 & 78.7 & - \\
        Rank\cite{rankedlist}(Iv2) & 512 & 57.4 & 69.7 & 79.2 & 62.6 \\
        HTG\cite{htg} & - & 59.5 & 71.8 & 81.3 & - \\
        Margin\cite{margin} & 128 & 63.6 & 74.4 & 83.1 &69.0 \\
        D\&C\cite{Sanakoyeu_2019_CVPR} & 128 & 65.9 & 76.6 & 84.4 &69.6 \\
        MIC\cite{mic} & 128 & 66.1 & 76.8 & 85.6 & 69.7\\
        \hline
        A-BIER\cite{abier} & 512 & 57.5 & 68.7 & 78.3& - \\
        Rank\cite{rankedlist}[Iv2] & 1536 & 61.3 & 72.7 & 82.7 & 66.1 \\
        DREML\cite{dreml} & 9216 & 63.9 & 75.0 & 83.1 &  67.8 \\
        \hline
        Margin(ReImp) & 256 & 65.2 & 75.9 & 84.5 & 68.1 \\
        Ours ($\mathcal{T}_{\mathcal{X}}^{\mathcal{G}_l}$)\cite{mic} & 256 & 67.0 & 77.3 & 85.8 & 69.3 \\
        \textbf{Ours ($\mathcal{T}_{\mathcal{X}}^*)$} & 256 & \textbf{68.6}& \textbf{79.4} & \textbf{86.8} & \textbf{71.0} \\
        \bottomrule
      \end{tabular}
\end{table}

%% file: tables_new/eval_sop.tex
\begin{table}[h!]
  \caption{Evaluation on SOP\cite{lifted}.}
  \label{tab:sop}
  \setlength\tabcolsep{1.4pt}
  \centering
  \begin{tabular}{l|c|ccc|c}
    \toprule
    Approach & Dim & R@1 & R@10 & R@100 & NMI \\
    \midrule
    HDML\cite{hardness-aware} & 512 & 68.7 & 83.2 & 92.4 & 89.3 \\
    DVML\cite{dvml} & 512 & 70.2 & 85.2 & 93.8 & \textbf{90.8} \\
    Margin\cite{margin} & 128 & 72.7 & 86.2 & 93.8 & 90.7 \\
    A-BIER\cite{abier} & 512 & 74.2 & 86.9 & 94.0 & - \\
    HTL\cite{htl} & 512 & 74.8 & 88.3 & 94.8 &  - \\
    D\&C\cite{Sanakoyeu_2019_CVPR} & 128 & 75.9 & 88.4 & 94.9 &  90.2 \\
    Rank\cite{rankedlist}[Iv2] & 512 & 76.1 & 89.1 & 95.4 & 89.7 \\
    MIC\cite{mic} & 128 & 77.2 & 89.4 & 95.6 & 90.0\\
    \hline
    Rank\cite{rankedlist}[Iv2] & 1536 & 79.8 & 91.3 & 96.3 & 90.4 \\
    \hline
    Margin(ReImp) & 256 & 76.1 & 88.1 & 94.9 & 89.5 \\
    Ours ($\mathcal{T}_{\mathcal{X}}^{\mathcal{G}_l}$)\cite{mic} & 256 & 77.7 & 89.8 & 95.9 & 90.0 \\
    \textbf{Ours ($\mathcal{T}_{\mathcal{X}}^*$)} & 256 & \textbf{78.2} & \textbf{90.1} & \textbf{96.1} & 90.3 \\
    \bottomrule
  \end{tabular}
\end{table}

%% file: tables_new/cub_cars_sop_grid.tex
\begin{table*}[h!]
  \caption{\textit{Evaluation using different ranking losses using ResNet50~\cite{resnet} backbone architecture.}
  Original: results reported in the original paper. Baseline$<$dim$>$: our implementation with ResNet50 and $<$dim$>$ embedding dimensions. $\phi$,$\phi^*$: class and shared embedding with 128 dimensions each. $\phi$+$\phi^*$: embedding concatenation resulting in 256 dimensions. 
  \textbf{bold}: best result for the given loss. \best{\textbf{underlined}}: best result on the dataset. We report 5-run average and standard deviation. 
  }
  \label{tab:grid_resnet}
  \setlength\tabcolsep{4.5pt}
  \centering
  \begin{tabular}{l|l|ccc|ccc}
    \toprule
    Approach & & Original & Baseline128 & Baseline256&$\phi$(Ours) & $\phi^*$(Ours) &  $\phi$+$\phi^*$(Ours) \\
    \hline\hline

    & \multicolumn{7}{c}{Dataset: CARS196\cite{cars196}} \\
    \hline
    Semihard\cite{semihard}  & R@1 & 51.5 & $71.9\pm0.3$ & $72.7\pm0.3$ & $72.7\pm0.6$ & $76.5\pm0.4$ & $\mathbf{79.4\pm0.3}$\\
                             & NMI & 53.4& $64.1\pm0.3$ & $64.5\pm0.4$ & $64.1\pm0.3$ & $63.2\pm0.3$ & $\mathbf{66.0\pm0.2}$\\
    \hline
    N-pairs\cite{npairs}     & R@1 & 71.1 & $70.2\pm0.3$& $70.6\pm0.2$  & $70.0\pm0.5$ & $74.8\pm0.5$ & $\mathbf{77.2\pm0.4}$\\
                             & NMI & 64.0 & $62.5\pm0.3$ & $62.3\pm0.2$ & $63.2\pm0.3$ & $62.7\pm0.1$ &  $\mathbf{64.3\pm0.1}$\\
    \hline
    PNCA\cite{proxynca}      & R@1 & 73.2 & $79.8\pm0.1$& $80.8\pm0.3$  & $80.2\pm0.1$ & $81.6\pm0.1$& $\mathbf{82.7\pm0.1}$\\
                             & NMI & 64.9 & $65.9\pm0.2$ &  $\mathbf{66.9\pm0.3}$ & $66.1\pm0.1$ & $64.5\pm0.3$ & $66.3\pm0.3$\\                        
    \hline

     Margin\cite{margin}      & R@1 & 79.6 & $80.1\pm0.2$& $81.5\pm0.3$ & $82.1\pm0.2$ & $86.2\pm0.2$ & \best{$\mathbf{87.0\pm0.1}$} \\
                             & NMI & 69.1 & $66.6\pm0.3$& $67.4\pm0.1$ & $68.3\pm0.3$ & $67.3\pm0.2$& \best{$\mathbf{69.8\pm0.1}$}\\
    \hline\hline
     & \multicolumn{7}{c}{Dataset: CUB200-2011\cite{cub200-2011}} \\
    \hline
    Semihard\cite{semihard}   & R@1 & 42.6 & $60.6\pm0.2$ & $61.7\pm0.3$&$60.2\pm0.1$&$62.5\pm0.2$&$\mathbf{64.6\pm0.1}$\\
                              & NMI & 55.4 & $65.5\pm0.3$ & $66.1\pm0.2$ & $65.8\pm0.3$ & $67.5\pm0.2$ & $\mathbf{68.5\pm0.1}$\\
    \hline
    N-pairs\cite{npairs}      & R@1 & 51.0 & $60.4\pm0.4$ & $60.2\pm0.3$  &$58.8\pm0.3$& $61.1\pm0.3$& $\mathbf{62.9\pm0.2}$\\
                              & NMI & 60.4 & $66.1\pm0.4$ & $65.0\pm0.2$&$65.1\pm0.4$& $66.0\pm0.2$&  $\mathbf{67.8\pm0.2}$\\
    \hline
    PNCA\cite{proxynca}       & R@1 & 61.9 & $64.0\pm0.1$ & $64.6\pm0.2$  &$63.6\pm0.2$& $65.7\pm0.2$& $\mathbf{66.4\pm0.2}$\\
                              & NMI & 59.5 & $\mathbf{68.1\pm0.2}$ & $68.0\pm0.2$&$67.5\pm0.2$& $67.5\pm0.3$&  $\mathbf{68.1\pm0.2}$\\    
    \hline
    Margin\cite{margin}       & R@1 & 63.6 & $63.6\pm0.3$ & $65.2\pm0.3$&$66.2\pm0.3$ & $67.4\pm0.4$ & \best{$\mathbf{68.6\pm0.2}$} \\
                              & NMI & 69.0 & $68.5\pm0.3$ & $68.1\pm0.3$& $69.2\pm0.5$& $69.7\pm0.6$& \best{$\mathbf{71.0\pm0.6}$}\\
    \hline\hline
     & \multicolumn{7}{c}{Dataset: SOP\cite{lifted}} \\
    \hline
    Semihard\cite{semihard}  & R@1 & -    & $73.5\pm0.2$ & $74.7\pm0.3$ & $75.3\pm0.2$ & $65.8\pm0.3$ & $\mathbf{75.5\pm0.2}$\\
                             & NMI & -    & $89.2\pm0.2$ & $89.4\pm0.2$ & $89.7\pm0.1$ & $86.9\pm0.2$ & $\mathbf{89.8\pm0.1}$\\
    \hline
    N-Pairs\cite{npairs}     & R@1 & 67.7 & $71.3\pm0.3$& $72.8\pm0.2$  & $74.1\pm0.2$ & $67.8\pm0.2$& $\mathbf{74.6\pm0.1}$\\
                             & NMI & 88.1 & $89.2\pm0.2$ &  $89.2\pm0.3$ & $89.8\pm0.2$ & $87.3\pm0.1$ & $\mathbf{89.9\pm0.1}$\\
    \hline
    Margin\cite{margin}      & R@1 & 72.7 & $74.4\pm0.2$& $76.1\pm0.3$ & $77.7\pm0.2$ & $72.2\pm0.2$ & \best{$\mathbf{78.2\pm0.1}$} \\
                             & NMI & 90.7 & $89.6\pm0.2$& $89.5\pm0.2$& $90.1\pm0.2$& $88.8\pm0.2$& \best{$\mathbf{90.3\pm0.1}$}\\
    \bottomrule
  \end{tabular}
\end{table*}

%% file: tables_new/analysis_gen_gap.tex
\begin{table}
    \caption{\textit{Generalization gap study}. The generalization gap is measured as the difference of performances on the train and test set of CARS196\cite{cars196} dateset. Performance measured in Recall@1. $\mathcal{N}(.)$ indicates random weight reset of the embedding layer connected to the feature extractor $f$. \textit{Dim.} denotes the dimensionalitay of a representation.}
    \label{tab:analysis_generalization}
    \setlength\tabcolsep{1.5pt}
    \centering
    \begin{tabular}{l|c|c|cc|c}
        \toprule
        Model & Representation & Dim. & Trainset & Testset & Generalization gap\\
        \midrule
         & $\phi$ & 128 & 90.7 &  79.9 & -10.8\\
         Margin\cite{margin} & $\phi$ ($\mathcal{N}$) & 128 & 87.5 & 81.4 & -6.1\\
         & $f$ & 2048 & 87.0 & 82.8 & -4.2\\
        \midrule
          & $\phi$ & 128 & 87.9 &  82.1 & -5.8\\
        Ours ($\mathcal{T}_{\mathcal{X}}^*$) & $\phi^*$ & 128 & 84.8 &  \textbf{86.2} & \textbf{1.4} \\
          & $\phi$ ($\mathcal{N}$) & 128 & 84.2 & 82.9 & -1.3\\
         & $f$ & 2048 & 83.7 & 83.5 & -0.2 \\
        \bottomrule
    \end{tabular}
\end{table}


%% file: tables_new/ablation_generalization_techniques.tex
\begin{table}
  \caption{\textit{Comparison of our approach against standard generalization techniques}. ResNet50\cite{resnet} is trained on CARS196\cite{cars196} and Recall@1\cite{recall} is reported. Embedding dimensionality is 128 for all cases.}
  \label{tab:generalization_noise}
  \setlength\tabcolsep{2.5pt}
  \centering
  \begin{tabular}{l|c|c|c|c||c}
    \toprule
      Noise & Margin\cite{margin} & + dropout & + noise output & + noise input & Ours\\

    \midrule
    Rec@1 & 79.9 & 80.5 & 79.8 & 80.4 & \textbf{83.2}\\ 
    \bottomrule
  \end{tabular}
\end{table}

%% file: tables_new/ablation_architecture_v2.tex
\begin{table}
    \caption{\textit{Architecture study} for our approach by computing Recall@1 on CARS196\cite{cars196} dataset. We compare the purely discriminative baseline leveraging only $\mathcal{T}_{\mathcal{X}}$ (\textit{discr. only}), a model trained only on shared characteristics, i.e. using only $\mathcal{T}_{\mathcal{X}}^*$ (\textit{shared only}), as well as learning both tasks simultaneously using the same encoder (\textit{both}), separate encoders (\textit{both+sep}) and separate encoders with decorrelation (\textit{both+sep+decor}). We further examine the impact of de-correlation weight $\gamma$.}
    \label{tab:ablation_architecture}
    \setlength\tabcolsep{1.5pt}
    \centering
    \begin{subtable}{0.45\textwidth}
    \centering
    \caption{Architectural Setups}
    \begin{tabular}{c||c|c|c|c|c}
        \toprule
        Arch. & discr. only\cite{margin} & shared only & both & both+sep & both+sep+decor\\
        Rec@1 & 81.5 & 36.1 & 83.2 & 84.1 & \textbf{87.0} \\
        \bottomrule
    \end{tabular}
    \end{subtable}
    \begin{subtable}{0.45\textwidth}
    \centering
    \vspace{7pt}
    \caption{Controlling the influence of decorrelation by varying the weighting $\gamma$ in the both+sep+decor setup.}
    \begin{tabular}{c||c|c|c|c|c|c|c}
        \toprule
        $\gamma$ &  0 & 5 & 50 & 200 & 500 & 1000 & 2500\\
        Rec@1 & 84.1 & 84.3 & 85.2 & 86.0 & \textbf{87.0} & 85.5 & 82.1\\
        \bottomrule
    \end{tabular}
    \end{subtable}    
\end{table}

%% file: tables_new/ablation_sampling_shared_v2.tex

\begin{table}
    \caption{\textit{Comparison of different sampling strategies and shared triplet setups to learn shared features}. Evaluation based on Recall@1 on the CARS196\cite{cars196} dataset.}
    \label{tab:ablation_sampling}
    \setlength\tabcolsep{1.5pt}
    \centering
    \begin{subtable}{0.45\textwidth}
    \centering
    \caption{Influence of various shared triplet setups.}    
    \begin{tabular}{l||c|c|c|c|c|c}
        \toprule
        Setup & Base\cite{margin} & $\mathcal{T}_{\mathcal{X}}^{\mathcal{G}_l}$ - std. & $\mathcal{T}_{\mathcal{X}}^{\mathcal{G}_l}$ + std. &  $\text{min}\;d(a,p)$ & NoConstr & $\mathcal{T}_{\mathcal{X}}^*$ \\
        Rec@1 & 81.5 & 82.3 & 85.8 & 82.1 &  86.0 & \textbf{87.0}\\
        \bottomrule
    \end{tabular}
    \end{subtable}
    \begin{subtable}{0.45\textwidth}
    \vspace{7pt}
    \centering
    \caption{Relevance of sampling methods for shared triplets. $\mathcal{T}_{\mathcal{X}}^*$ uses distance-based sampling \cite{margin} by default. $\phi$ is trained with \cite{margin} for each test. Only sampling for $\phi^*$ is changed.}    
    \begin{tabular}{l||c|c|c}
        \toprule
        Sampling & $\mathcal{T}_{\mathcal{X}}^*$ + random& $\mathcal{T}_{\mathcal{X}}^*$ + semihard\cite{semihard}&
        $\mathcal{T}_{\mathcal{X}}^*$ (distance \cite{margin})\\
        Rec@1 & 82.0 & 85.3 & \textbf{87.0}\\
        \bottomrule
    \end{tabular}
    \end{subtable}    
\end{table}

%% file: tables_new/eval_dml_base_inceptionBN.tex
\begin{table*}[h!]
  \caption{\textit{Evaluation using different ranking losses with GoogLeNet\cite{googlenet} backbone architecture.}
  Original: results reported in the original paper. Note that the original works of Margin\cite{margin} uses ResNet50 and ProxyNCA (PNCA)\cite{proxynca} uses Inception-BN. All other use GoogLeNet. Baseline$<$dim$>$: our re-implementation with GoogLeNet architecture and $<$dim$>$ embedding dimensions. $\phi$,$\phi^*$: class and shared embedding with 512 dimensions each. $\phi$+$\phi^*$: embedding concatenation resulting in 1024 dimensions.   \textbf{bold}: best result for the given loss. \best{\textbf{underlined}}: best result on the dataset. We report 5-run average and standard deviation. 
  }

  \label{tab:grid_ibn}
  \setlength\tabcolsep{4.5pt}
  \centering
  \begin{tabular}{l|l|ccc|ccc}
    \toprule
    Approach & & Original & Baseline512 & Baseline1024&$\phi$(Ours) & $\phi^*$(Ours) &  $\phi$+$\phi^*$(Ours) \\
    \hline\hline
     & \multicolumn{7}{c}{Dataset: CARS196\cite{cars196}} \\
    \hline
    Semihard\cite{semihard}& R@1 & 51.5 & $67.3\pm0.3$ & $66.3\pm0.4$ & $64.4\pm0.6$ & $71.4\pm0.2$ & $\mathbf{72.7\pm0.2}$\\
                           & NMI & 53.4 & $60.3\pm0.3$ & $58.9\pm0.2$ & $56.9\pm0.2$ & $58.2\pm0.5$ & $\mathbf{60.5\pm0.2}$\\
    \hline
    N-pairs\cite{npairs}   & R@1 & 71.1 & $67.4\pm0.4$ & $64.5\pm1.2$ & $60.4\pm0.8$ & $67.8\pm0.3$ & $\mathbf{67.0\pm0.2}$\\
                           & NMI & 64.0 & $60.0\pm0.3$ & $58.6\pm0.4$ & $56.1\pm0.2$ & $59.5\pm0.3$ & $\mathbf{60.1\pm0.2}$\\
    \hline
    PNCA\cite{proxynca}    & R@1 & 73.2 & $71.1\pm0.2$ & $71.2\pm0.2$ & $73.8\pm0.2$ & $76.7\pm0.2$ & \best{$\mathbf{78.3\pm0.1}$}\\
                           & NMI & 64.9 & $60.2\pm0.2$ & $58.9\pm0.2$ & $61.7\pm0.2$ & $62.3\pm0.4$ & $\mathbf{63.8\pm0.3}$\\                        
    \hline
    Margin\cite{margin}    & R@1 & 79.6 & $72.6\pm0.3$ & $73.3\pm0.2$ & $74.3\pm0.3$ & $75.5\pm0.3$ & $\mathbf{77.4\pm0.3}$ \\
                           & NMI & 69.1 & $63.2\pm0.2$ & $62.2\pm0.2$ & $63.3\pm0.5$ & $59.3\pm0.4$ & \best{$\mathbf{64.1\pm0.3}$}\\
    \hline\hline

     & \multicolumn{7}{c}{Dataset: CUB200-2011\cite{cub200-2011}} \\
    \hline
    Semihard\cite{semihard}& R@1 & 42.6 & $54.8\pm0.3$ & $56.6\pm0.4$ & $57.4\pm0.3$ & $60.2\pm0.3$ & $\mathbf{60.9\pm0.2}$\\
                           & NMI & 55.4 & $61.9\pm0.2$ & $62.5\pm0.5$ & $63.5\pm0.4$ & $65.3\pm0.3$ & $\mathbf{66.0\pm0.2}$\\
    \hline
    N-pairs\cite{npairs}   & R@1 & 51.0 & $52.2\pm0.3$ & $50.9\pm1.4$ & $50.8\pm0.1$ & $52.5\pm0.2$ & $\mathbf{54.8\pm0.2}$\\
                           & NMI & 60.4 & $60.7\pm0.3$ & $59.2\pm1.4$ & $59.5\pm0.3$ & $61.1\pm0.3$ & $\mathbf{61.6\pm0.2}$\\
    \hline
    PNCA\cite{proxynca}    & R@1 & 61.9 & $55.9\pm0.2$ & $56.4\pm0.3$ & $58.6\pm0.2$ & $60.8\pm0.1$ & $\mathbf{61.5\pm0.1}$\\
                           & NMI & 59.5 & $62.9\pm0.1$ & $62.0\pm0.2$ & $64.5\pm0.2$ & $65.5\pm0.2$ &  $\mathbf{65.8\pm0.2}$\\    
    \hline
    Margin\cite{margin}    & R@1 & 63.6 & $58.3\pm0.3$ & $59.3\pm0.3$ & $60.9\pm0.3$ & $61.9\pm0.3$ & \best{$\mathbf{62.6\pm0.2}$} \\
                           & NMI & 69.0 & $64.8\pm0.2$ & $64.4\pm0.3$ & $65.0\pm0.3$ & $66.2\pm0.2$ & \best{$\mathbf{66.7\pm0.2}$}\\
    \hline\hline
     & \multicolumn{7}{c}{Dataset: SOP\cite{lifted}} \\
    \hline
    Semihard\cite{semihard}&R@1 & -     & $67.3\pm0.1$ & $67.4\pm0.3$ & $70.7\pm0.2$ & $67.5\pm0.3$ & $\mathbf{71.1\pm0.2}$\\
                           &NMI & -     & $88.4\pm0.1$ & $88.4\pm0.3$ & $88.6\pm0.1$ & $86.8\pm0.1$ & $\mathbf{89.2\pm0.1}$\\
    \hline
    N-Pairs\cite{npairs}   &R@1 & 67.7  & $67.2\pm0.3$ & $63.4\pm0.4$ & $68.3\pm0.3$ & $66.6\pm0.1$ & $\mathbf{68.9\pm0.1}$\\
                           &NMI & 88.1  & $88.3\pm0.2$ & $87.2\pm0.3$ & $88.5\pm0.2$ & $86.4\pm0.3$ & $\mathbf{88.8\pm0.1}$\\
    \hline
    Margin\cite{margin}    &R@1 & 72.7  & $68.5\pm0.2$ & $67.1\pm0.3$ & $71.0\pm0.3$ & $69.2\pm0.3$ & \best{$\mathbf{72.0\pm0.3}$} \\
                           &NMI & 90.7  & $88.6\pm0.2$ & $87.6\pm0.3$ & $88.4\pm0.1$ & $87.5\pm0.2$ & \best{$\mathbf{89.1\pm0.1}$}\\
    \bottomrule
  \end{tabular}
\end{table*}